\documentclass[10pt,conference,a4paper]{IEEEtran}
\usepackage[utf8]{vntex}
\usepackage{amstext,amsmath,latexsym,amsbsy,amssymb}

\usepackage[rightcaption,raggedright]{sidecap}
\usepackage{framed} 
\usepackage{soul} 
\usepackage{rotating}
\usepackage{floatpag}
	\rotfloatpagestyle{empty}
\usepackage{amsthm}
\usepackage{graphicx, longtable}
\usepackage{makeidx}    
    \makeindex
\let\labelindent\relax

\usepackage{fancyvrb}   
\usepackage{fvextra}    

\usepackage{pdflscape}
\usepackage{amsfonts}
\usepackage{url}
\usepackage[printonlyused]{acronym}
	
\usepackage{tabularx,multicol,multirow,longtable}
\usepackage{makecell}
\usepackage{color,colortbl}
\usepackage{exscale,eucal}
\usepackage{steinmetz}
\usepackage{subcaption}
\usepackage{enumitem}
\usepackage{cite}
\usepackage{lipsum}
\usepackage{algpseudocode,algorithm}
    \makeatletter
    \renewcommand{\ALG@name}{Thuật toán}
    \makeatother
\usepackage{color} 
\usepackage{array}
	\newcolumntype{P}[1]{>{\raggedright\arraybackslash}p{#1}}

\usepackage{listings}
\usepackage{balance}
\usepackage{tikz}
	\usetikzlibrary{calc,arrows,decorations.pathmorphing,backgrounds,positioning,fit,shapes,decorations.shapes, shapes.geometric, decorations.pathreplacing, decorations.text}
\usepackage{circuitikz}

\usepackage[update]{epstopdf}
\DeclareGraphicsExtensions{%
    .png,.PNG,%
    .pdf,.PDF,%
    .jpg,.mps,.jpeg,.jbig2,.jb2,.JPG,.JPEG,.JBIG2,.JB2}

\usepackage[hidelinks]{hyperref}

\usepackage{nameref} 

\usepackage{adjustbox}
\usepackage{fp}

\tikzset{
    line/.style={draw, thick},
}

\tikzset{XOR/.style={draw,circle,append after command={
        [shorten >=\pgflinewidth, shorten <=\pgflinewidth,]
        (\tikzlastnode.north) edge (\tikzlastnode.south)
        (\tikzlastnode.east) edge (\tikzlastnode.west)
        }
    }
}

\tikzset{decorate sep/.style 2 args=
{decorate,decoration={shape backgrounds,shape=circle,shape size=#1,shape sep=#2}}}

\tikzstyle{startstop} = [rectangle, rounded corners, minimum width=3cm, minimum height=1cm,text centered, draw=black, fill=red!10!white]

\tikzstyle{io} = [trapezium, trapezium left angle=70, trapezium right angle=110, minimum width=3cm, minimum height=1cm, text centered, draw=black, fill=orange!10!white]

\tikzstyle{process} = [rectangle, minimum width=3cm, minimum height=1cm, text centered, draw=black, fill=blue!10!white]

\tikzstyle{decision} = [diamond, minimum width=3cm, minimum height=1cm, text centered, draw=black, fill=green!10!white]

\tikzstyle{arrow} = [thick,->,>=stealth]

\tikzstyle{darrow} = [thick, stealth-stealth]

\tikzstyle{arrowt1}  = [thick,->,>=stealth, draw=red]

\tikzstyle{arrowt2}  = [thick,->,>=stealth, dashed, draw=green]

\tikzstyle{arrowt3}  = [thick,->,>=stealth, dotted, draw=black]

\tikzstyle{field} = [rectangle, minimum width=5mm, minimum height=10mm, align=center, fill=green!10!white, font=\footnotesize, draw=black, line width=1pt]

\tikzstyle{field1} = [rectangle, minimum width=20mm, minimum height=5mm, align=center, fill=green!10!white, font=\scriptsize, draw=black, line width=1pt, rotate=-90]

\tikzstyle{field2} = [rectangle, minimum width=5mm, minimum height=20mm, align=center, fill=green!10!white, font=\scriptsize, draw=black, line width=1pt]

\tikzstyle{field3} = [rectangle, minimum width=31mm, minimum height=5mm, align=center, fill=green!10!white, font=\small, draw=black, line width=1pt, rotate=-90]

\tikzstyle{field4} = [rectangle, minimum width=31mm, minimum height=5mm, align=center, fill=green!10!white, draw=black, line width=1pt, rotate=-90]

\tikzstyle{field5} = [rectangle, minimum width=3mm, minimum height=10mm, align=center, fill=blue!10!white, draw=black, line width=1pt]

\tikzstyle{fieldx} = [rectangle, minimum width=3mm, minimum height=10mm, align=center, fill=green!10!white, draw=black, line width=1pt]

\tikzstyle{fieldxx} = [rectangle, minimum width=3mm, minimum height=10mm, align=center, fill=red!10!white, draw=black, line width=1pt]

\tikzstyle{field6} = [rectangle, minimum width=3mm, minimum height=10mm, align=center, fill=green!10!white, font=\small, draw=black, line width=1pt]

\tikzstyle{field7} = [rectangle, minimum width=3mm, minimum height=10mm, align=center, fill=green!10!white, font=\footnotesize, draw=black, line width=1pt]

\makeatletter
\tikzset{
    database/.style={
        path picture={
            \draw (0, 1.5*\database@segmentheight) circle [x radius=\database@radius,y radius=\database@aspectratio*\database@radius];
            \draw (-\database@radius, 0.5*\database@segmentheight) arc [start angle=180,end angle=360,x radius=\database@radius, y radius=\database@aspectratio*\database@radius];
            \draw (-\database@radius,-0.5*\database@segmentheight) arc [start angle=180,end angle=360,x radius=\database@radius, y radius=\database@aspectratio*\database@radius];
            \draw (-\database@radius,1.5*\database@segmentheight) -- ++(0,-3*\database@segmentheight) arc [start angle=180,end angle=360,x radius=\database@radius, y radius=\database@aspectratio*\database@radius] -- ++(0,3*\database@segmentheight);
        },
        minimum width=2*\database@radius + \pgflinewidth,
        minimum height=3*\database@segmentheight + 2*\database@aspectratio*\database@radius + \pgflinewidth,
    },
    database segment height/.store in=\database@segmentheight,
    database radius/.store in=\database@radius,
    database aspect ratio/.store in=\database@aspectratio,
    database segment height=0.1cm,
    database radius=0.25cm,
    database aspect ratio=0.35,
}
\makeatother

\begin{document}
\columnsep=0.63cm

%
\title{Ước lượng kênh truyền trong hệ thống đa robot\\sử dụng SDR}

\author{
\IEEEauthorblockN{
Đỗ Hải Sơn\IEEEauthorrefmark{1},
Nguyễn Hữu Hưng\IEEEauthorrefmark{2},
Phạm Duy Hưng\IEEEauthorrefmark{2},
Trần Thị Thúy Quỳnh\IEEEauthorrefmark{2}
} 
\IEEEauthorblockA{\IEEEauthorrefmark{1} Viện Công nghệ Thông tin - Đại học Quốc gia Hà Nội\\ \IEEEauthorrefmark{2} Trường Đại học Công nghệ - Đại học Quốc gia Hà Nội\\ 
		Email: dohaison1998@vnu.edu.vn, nguyenhuuhung032@gmail.com, \{hungpd, quynhttt\}@vnu.edu.vn}
}
\maketitle

\begin{abstract}
Nghiên cứu này tập trung xây dựng một hệ thống thực nghiệm ước lượng kênh truyền phục vụ truyền thông trong hệ thống đa robot di động sử dụng các thiết bị vô tuyến định nghĩa bằng phần mềm (SDR - Software Define Radio). Hệ thống gồm 2 robot di động được lập trình cho 2 tình huống robot đứng yên và robot di động theo quỹ đạo cho trước. Hệ thống truyền thông được thực hiện thông qua điều chế đa sóng mang trực giao OFDM (Orthogonal Frequency Division Multiplexing) nhằm làm giảm ảnh hưởng của môi trường đa đường trong nhà. Hiệu năng của hệ thống được đánh giá thông qua tỷ lệ lỗi bit BER (Bit Error Rate). Các kết nối liên quan đến chuyển động của robot và phần truyền thông được thực hiện tương ứng thông qua Raspberry Pi 3 và BladeRF x115. Kỹ thuật bình phương tối thiểu LS (Least Squares) được thực hiện để ước lượng kênh với tỷ lệ lỗi bit khoảng $10^{-2}$.
	 
\end{abstract}

\IEEEpeerreviewmaketitle

\section{GIỚI THIỆU}
Mạng truyền thông trong các hệ thống đa robot đang là chủ đề được quan tâm hiện nay~\cite{Chen2021} do gắn liền với cách mạng công nghiệp 4.0. Trong các mạng này, mỗi robot thường yêu cầu có các bộ thu phát vô tuyến để giao tiếp với các robot khác trong mạng hoặc máy chủ điều khiển từ bên ngoài. Tuy nhiên, trong các điều kiện môi trường phức tạp~\cite{Girma2020}, các robot cỡ nhỏ sẽ gặp phải khó khăn khi phải ước lượng kênh truyền vô tuyến để thiết lập tuyến truyền thông chính xác. Một số phương pháp ước lượng kênh truyền cho các robot cỡ nhỏ đã được đề cập trong~\cite{Burgard2008, Gao2023}. Tuy nhiên, các nghiên cứu thực nghiệm về vấn đề này còn khá mới hoặc các sản phẩm thương mại thường sử dụng các bo mạch có sẵn, khó can thiệp. 

Trong những năm gần đây, các thiết bị vô tuyến định nghĩa bằng phần mềm SDR (Software Define Radio) được sử dụng nhiều cho mục đích kiểm nghiệm các thuật toán trong viễn thông~\cite{nc}, do tính linh hoạt và đa dạng trong các thư viện xử lý tín hiệu có sẵn trên phần mềm GNU Radio~\cite{gnuradio}. Từ kết quả trong nghiên cứu trước đây của nhóm~\cite{REV2021}, các thiết bị SDR có khả năng triển khai trên các robot cỡ nhỏ với nguồn điện độc lập được cấp từ pin ngay trên robot. Trong nghiên cứu này, một hệ thống thử nghiệm gồm hai robot cỡ nhỏ được điều khiển bằng Raspberry Pi 3~\cite{pi3}, mỗi robot lắp một thiết bị SDR (BladeRF x115~\cite{nuand}) để thực hiện việc thu phát và ước lượng kênh truyền thời gian thực. Hệ thu phát sử dụng điều chế ghép kênh phân chia theo tần số trực giao OFDM (Orthogonal Frequency-division Multiplexing) với mã nguồn mở có tên SDR4All tương tự như chuẩn WiFi 802.11 được phát triển bởi~\cite{sdrnc,Ejder2012,nc}. Mã nguồn mở này sử dụng bộ cân bằng ép về không ZF (Zero Forcing) để ước lượng kênh truyền vô tuyến. Trong bài báo này, nhóm nghiên cứu không thay đổi thuật toán bình phương tối thiểu LS (Least Squares) của SDR4All mà tập trung vào xem xét hiệu năng của giải thuật này trên các robot trong các điều kiện khác nhau. Hiệu năng của hệ thống được đánh giá thông qua tỷ lệ lỗi bit BER (Bit Error Rate).

Các đóng góp chính của bài báo gồm: xây dựng các khối trên GNU Radio dùng để mô phỏng kênh truyền vô tuyến và tính hệ số BER; đánh giá chất lượng của ước lượng kênh truyền trong hệ thống đa robot thông qua mô phỏng và kiểm nghiệm thực tế.

Bài báo được tổ chức như sau: phần~\ref{Sec:MoHinhHeThong} trình bày về mô hình hệ thống, sơ lược về lý thuyết sử dụng cho ước lượng kênh truyền trong bộ công cụ SDR4All. Kết quả mô phỏng và thực nghiệm được biểu diễn trong phần~\ref{Sec:KetQuaMoPhong}. Một vài thảo luận về hiệu năng của hệ thống sau khi thực nghiệm được trình bày trong phần~\ref{Sec:dis}. Kết luận của bài báo được đưa ra trong phần~\ref{Sec:KetLuan}.
\section{ƯỚC LƯỢNG KÊNH TRUYỀN TRONG SDR4All}
\label{Sec:MoHinhHeThong}

Phần này trình bày về mô hình thu phát sử dụng mã nguồn SDR4ALL, bao gồm cấu trúc của các khung OFDM và giải thuật của bộ cân bằng ZF.


\begin{figure*} [t]
	\centering
    \begin{tikzpicture}
        \node[rectangle, draw=black, dashed, fill=black!5!white, minimum width=73mm, minimum height=35mm] at (45mm, 0) {};
        \node[] (PC1)
            {\includegraphics[width=.07\textwidth]{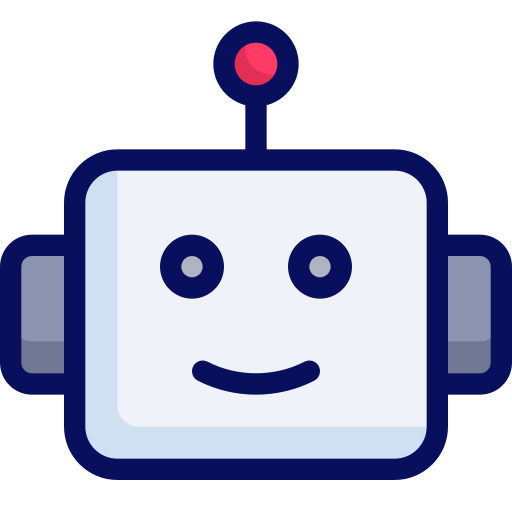}};
        \node[below=1mm of PC1] (R1) {Robot 1};
        \node[right=3mm of PC1, field5, minimum width=20mm] (B11) {Raspberry\\Pi 3: \\ GNU Radio};
        \node[right=30mm of PC1, field5, minimum width=20mm] (B12) {BladeRF 1};

        \node[rectangle, draw=black, dashed, fill=black!5!white, minimum width=73mm, minimum height=35mm] at (122mm, 0) {};
        
        \node[right=50mm of B12, field5, minimum width=20mm] (B22) {BladeRF 2};
        \node[right=7mm of B22, field5, minimum width=20mm] (B23) {Raspberry\\Pi 3: \\ GNU Radio};
        \node[] (PC2) at (167mm, 0)
            {\includegraphics[width=.07\textwidth]{figures/robot.png}};
        \node[below=1mm of PC2] (R2) {Robot 2};

        \node[antenna, thick, scale=0.5, label={left, font=\small}:{Ăng-ten 2}] at ([xshift=45mm, yshift=0mm]B12.east) (A2) {};

        \node[antenna, thick, scale=0.5, label={right, font=\small}:{Ăng-ten 1}] at ([xshift=6mm, yshift=0mm]B12.east) (A1) {};

        \draw[line] (B12) -- (A1);
        \draw[arrow] (B11) -- (B12);

        \draw[line] (B22) -- (A2);
        \draw[arrow] (B23) -- (B22);

        \draw[arrowt3] ([xshift=2mm, yshift=5mm]A1.east) -- ([xshift=-2mm, yshift=5mm]A2.west);
    \end{tikzpicture}
	\caption{Mô hình hệ thống thực nghiệm ước lượng kênh truyền cho robot sử dụng SDR.}
	\label{fig:SM}
\end{figure*}

\subsection{Mô hình hệ thống}
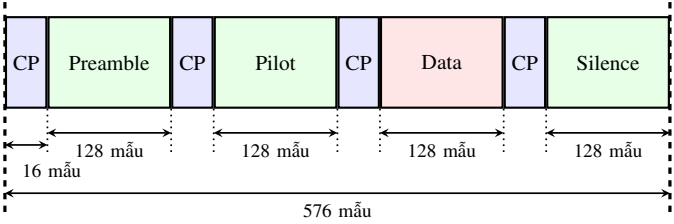
\begin{figure}[!t]
\begin{center}
    \scalebox{.8} {
    \begin{tikzpicture}[node distance = 0mm]
        \node[field5, minimum height=15mm] (B11) {CP};
        \node[right=0mm of B11, fieldx, minimum height=15mm, minimum width=20mm] (B12) {Preamble};
        \node[right=0mm of B12, field5, minimum height=15mm] (B13) {CP};
        \node[right=0mm of B13, fieldx, minimum height=15mm, minimum width=20mm] (B14) {Pilot};
        \node[right=0mm of B14, field5, minimum height=15mm] (B15) {CP};
        \node[right=0mm of B15, fieldxx, minimum height=15mm, minimum width=20mm] (B16) {Data};
        \node[right=0mm of B16, field5, minimum height=15mm] (B17) {CP};
        \node[right=0mm of B17, fieldx, minimum height=15mm, minimum width=20mm] (B18) {Silence};
        
        \draw[line, line width=1.5pt, dashed] ([yshift=10mm]B11.west) -- ([yshift=-25mm]B11.west);
        \draw[line, line width=1.5pt, dashed] ([yshift=10mm]B18.east) -- ([yshift=-25mm]B18.east);
        
        \draw[line, thick, dotted] (B11.south east) -- ([yshift=-15mm]B11.east);
        \draw[line, thick, dotted] (B12.south east) -- ([yshift=-15mm]B12.east);
        \draw[line, thick, dotted] (B13.south east) -- ([yshift=-15mm]B13.east);
        \draw[line, thick, dotted] (B14.south east) -- ([yshift=-15mm]B14.east);
        \draw[line, thick, dotted] (B15.south east) -- ([yshift=-15mm]B15.east);
        \draw[line, thick, dotted] (B16.south east) -- ([yshift=-15mm]B16.east);
        \draw[line, thick, dotted] (B17.south east) -- ([yshift=-15mm]B17.east);
        
        \draw[darrow] ([yshift=-4mm]B12.south west) -- ([yshift=-4mm]B12.south east) node [pos=0.5, below, font=\small] {128 mẫu};
        \draw[darrow] ([yshift=-4mm]B14.south west) -- ([yshift=-4mm]B14.south east) node [pos=0.5, below, font=\small] {128 mẫu};
        \draw[darrow] ([yshift=-4mm]B16.south west) -- ([yshift=-4mm]B16.south east) node [pos=0.5, below, font=\small] {128 mẫu};
        \draw[darrow] ([yshift=-4mm]B18.south west) -- ([yshift=-4mm]B18.south east) node [pos=0.5, below, font=\small] {128 mẫu};
        
        \draw[darrow] ([yshift=-6mm]B11.south west) -- ([yshift=-6mm]B11.south east) node [pos=1.1, below=4pt, font=\small] {16 mẫu};
        
        \draw[darrow] ([yshift=-14mm]B11.south west) -- ([yshift=-14mm]B18.south east) node [pos=0.5, below, font=\small] {576 mẫu};
    \end{tikzpicture}
    }
 \caption{Minh họa cấu trúc một khung OFDM của mã nguồn SDR4All.}
 \label{fig:OFDM}
\end{center}
\end{figure}
Xét mô hình hệ thống gồm hai robot đơn ăng-ten thu phát như trên hình~\ref{fig:SM} sử dụng điều chế OFDM theo mã nguồn SDR4All. Trong đó, mỗi robot bao gồm các thành phần điều khiển, chuyển động, một máy tính cỡ nhỏ là Raspberry Pi 3 và một BladeRF x115. Raspberry Pi chịu trách nhiệm điều khiển robot và chạy GNU Radio để gửi và nhận dữ liệu từ BladeRF x115. Tất cả các linh kiện trên robot di động đều được cấp nguồn bằng một pin gắn ngoài ngay trên robot. Lý thuyết về việc truyền nhận OFDM đã được biết đến rộng rãi~\cite{Yong2010} nên trong bài báo này sẽ không nhắc lại mà tập trung vào giới thiệu tổ chức khung OFDM của mã nguồn SDR4All như trên hình~\ref{fig:OFDM}~\cite{TTCT2C2022}. Khung dữ liệu OFDM được tạo ra bao gồm 4 thành phần chính: Preamble, Pilot, Data, và Silence, được truyền liên tục theo thời gian. Preamble, Pilot, và Silence là các thành phần không mang thông tin, có tác dụng lần lượt là đồng bộ hệ thống phát, ước lượng kênh truyền, và ước lượng tỷ số SNR. Biểu diễn của Preamble  tại thời điểm $n$ như sau:
\begin{equation}
    \mathbf{g} = \big[g_{1}[n], g_{2}[n],~\ldots, g_{N_{Occ}}[n] \big]^\top,
\end{equation}
với $\mathbf{g} \in \mathbb{C}^{N_{Occ} \times 1}$ và $N_{Occ}$ là số sóng mang con được sử dụng~\cite{Wu2009}. Các giá trị $g[n]$ được định nghĩa trong~\cite{Schmidl1997}. Thành phần Pilot được chèn vào khung OFDM theo kiểu Block-type~\cite{Ladaycia2017} tức được chèn vào toàn bộ các tần số ở các khe thời gian xác định. Các thành phần Pilot, Data, và Silence lần lượt có kích thước $\mathbf{p} \in \mathbb{C}^{N_{Occ} \times 1}$, $\mathbf{D} \in \mathbb{C}^{N_{Occ} \times N_d}$, $\mathbf{z} \in \mathbb{C}^{N_{Occ} \times 1}$, với $N_d$ là độ dài véc-tơ dữ liệu truyền đi, và được biểu diễn như sau:
\begin{equation}
    \mathbf{p} = \big[p_{1}[n], p_{2}[n],~\ldots, p_{N_{Occ}}[n] \big]^\top.
\end{equation}
\begin{equation}
\mathbf{D}=\left[\begin{array}{ccc}
d_{1}[n] & \ldots & d_{1}\left[n+N_d-1\right] \\
d_{2}[n] & \ldots & d_{2}\left[n+N_d-1\right] \\
\vdots & \ddots & \vdots \\
d_{N_{Occ}}[n] & \ldots & d_{N_{Occ}}\left[n+N_d-1\right]
\end{array}\right].
\end{equation}
\begin{equation}
    \mathbf{z} = \big[z_{1}[n], z_{2}[n],~\ldots, z_{N_{Occ}}[n] \big]^\top.
\end{equation}

Một khung OFDM $\mathbf{S} \in \mathbb{C}^{N_{Occ} \times (3 + N_d)}$ chỉ chứa 4 thành phần kể trên được biểu diễn như dưới đây~\cite{Ejder2012}:
\begin{equation}
    \mathbf{S} = \big[\mathbf{g} \mid \mathbf{p} \mid \mathbf{D} \mid \mathbf{z}\big],
\end{equation}
Để đơn giản hóa ký hiệu, các sóng mang con trống ($N_l$) ở hai phía trái phải của phổ chưa được đề cập. Trên thực tế, $2N_l \times (3 + N_d)$ sóng mang con sẽ được chèn vào để tạo thành khung OFDM có kích thước $(N_{Occ} + 2 N_l)\times(3+N_d)$. Sau các khối IFFT và chuẩn hóa năng lượng, cả bốn thành phần kể trên sẽ được chèn thêm tiền tố vòng CP (Cyclic prefix) bằng cách sao chép $N_{CP}$ mẫu cuối cùng mỗi luồng song song rồi chèn lên đầu để tạo thành khoảng bảo vệ. Kích thước cuối cùng của khung OFDM được robot truyền đi như trên hình~\ref{fig:OFDM} là $(N_{Occ} + 2 N_l + N_{CP})\times(3+N_d)$.

\subsection{Bộ cân bằng kênh Zero Forcing}
Ở phía robot thu, sau khi qua các bước như đồng bộ, chuyển đổi nối tiếp thành song song, loại bỏ CP, và FFT sẽ tách dữ liệu nhận thành dạng các khung OFDM như bên phát. Chi tiết các bước này có thể xem tại~\cite{TTCT2C2022, Ejder2012}. Ma trận của khung OFDM nhận được bên thu là:
\begin{equation}
    \hat{\mathbf{S}} = [\hat{\mathbf{g}} \mid \hat{\mathbf{p}} \mid \hat{\mathbf{D}} \mid \hat{\mathbf{z}}].
\end{equation}

Dựa vào thành phần $\hat{\mathbf{p}}$, bộ cân bằng ZF ước lượng ảnh hưởng của kênh truyền đến các ký hiệu pilot biết trước qua đó đảo ngược quá trình này để khôi phục các tín hiệu dữ liệu $\mathbf{D}$. Ma trận kênh cân bằng ZF $\hat{\mathbf{H}}_{est} \in \mathbb{C}^{N_{Occ} \times N_{Occ}}$ được ước lượng bởi:
\begin{equation}
    \hat{\mathbf{H}}_{est} = \operatorname{diag} \Big( \big[
    \frac{1}{\hat{h}_{N_l}}, \frac{1}{\hat{h}_{N_l+1}},~\ldots, \frac{1}{\hat{h}_{N_l+N_{Occ}-1}}
     \big] \Big ),
\end{equation}
với
\begin{equation}
    \hat{h}[k] = \frac{\hat{p}[k]}{p[k]}, \quad N_l \le k \le N_l + N_{Occ} - 1,
\end{equation}

Dữ liệu (các sóng mang con chứa thông tin) được khôi phục như sau:
\begin{equation}
    \hat{\mathbf{D}}_{est} = \hat{\mathbf{H}}_{est} \hat{\mathbf{D}}.
\end{equation}

Ngoài ước lượng tín hiệu bên phát, tỷ số SNR cũng được tính toán sử dụng thành phần Preamble và Silence như trong~(\ref{eq:snr}).
\begin{equation}
    \label{eq:snr}
    \operatorname{\hat{SNR}} = 10 \log \Big(
    \frac{\sum_{k=N_l}^{k=N_l + N_{Occ} - 1} \mid \hat{g}_k \mid ^2}
    {\sum_{k=N_l}^{k=N_l + N_{Occ} - 1} \mid \hat{z}_k \mid ^2}
    \Big).
\end{equation}

Các bước còn lại như chuyển đổi luồng song song thành nối tiếp và giải ánh xạ chòm sao để khôi phục các bit gửi đi xem tại~\cite{TTCT2C2022}.
\section{KẾT QUẢ MÔ PHỎNG VÀ THỰC NGHIỆM}
\label{Sec:KetQuaMoPhong}
Phần này sẽ trình bày về các bước thiết lập mô phỏng và thực nghiệm. Phân tích kết quả được đưa ra cùng với đánh giá về hiệu năng của hệ thống trong các kịch bản thử nghiệm khác nhau.

\subsection{Kết quả mô phỏng}

\subsubsection{Thiết lập mô phỏng}

Từ lý thuyết đã trình bày ở mục~\ref{Sec:MoHinhHeThong}, trước hết, một mô phỏng trên Matlab được triển khai dựa trên~\cite{Yong2010}. Trong đó, kênh truyền (thành phần đa đường) sẽ được mô hình hóa dưới dạng một bộ lọc đáp ứng xung hữu hạn FIR (Finite Impulse Response). Các hệ số của bộ lọc FIR được lựa chọn ngẫu nhiên, trong nghiên cứu này, chúng tôi chọn một bộ lọc FIR có độ dài bằng 2 như sau:
\begin{equation}
    \label{eq:fir}
    \mathbf{h} \equiv \operatorname{FIR}\big([0,8 + 0,9i; 0,6 + 0,7i]\big).
\end{equation}

Đối với mô phỏng trên GNU Radio, bài báo này sử dụng phiên bản GNU Radio~\texttt{3.7.11}~\footnote{\url{https://github.com/DoHaiSon/SDR_NC/blob/master/Documents/Readme.pdf}}. Mã nguồn SDR4All được sử dụng để xây dựng một bộ thu phát sử dụng OFDM hoàn chỉnh. Ảnh hưởng của kênh truyền được thực hiện thông qua khối ``\texttt{FIR Channel}'' do nhóm nghiên cứu xây dựng.

\subsubsection{Kết quả}

\begin{figure}
     \centering
     \begin{subfigure}[b]{0.5\textwidth}
         \centering
         \includegraphics[width=\textwidth]{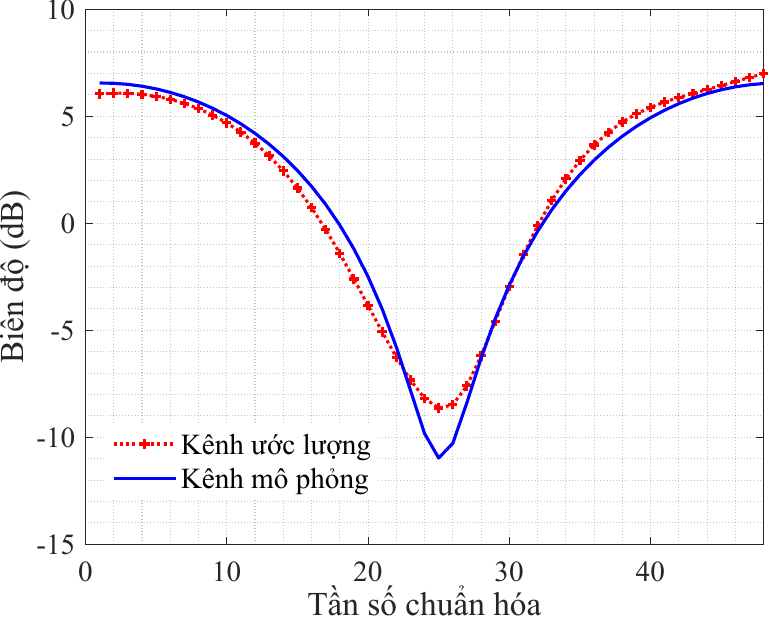}
         \caption{Matlab}
         \label{fig:simmatlab}
     \end{subfigure}
     \hfill
     \begin{subfigure}[b]{0.5\textwidth}
         \centering
         \includegraphics[width=.48\textwidth]{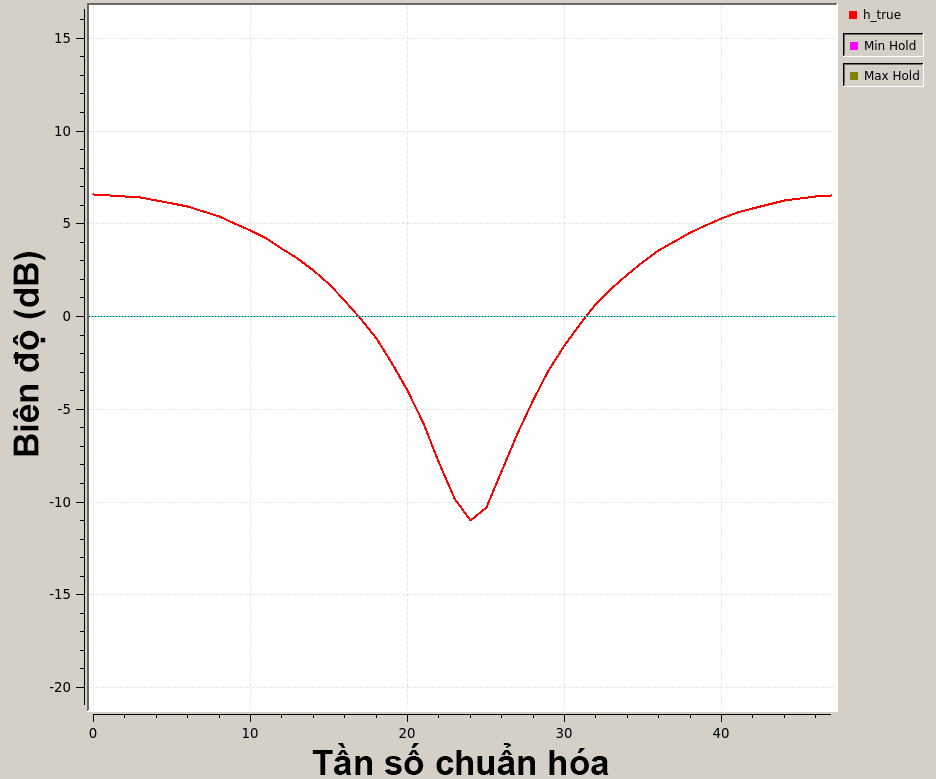}
         \includegraphics[width=.48\textwidth]{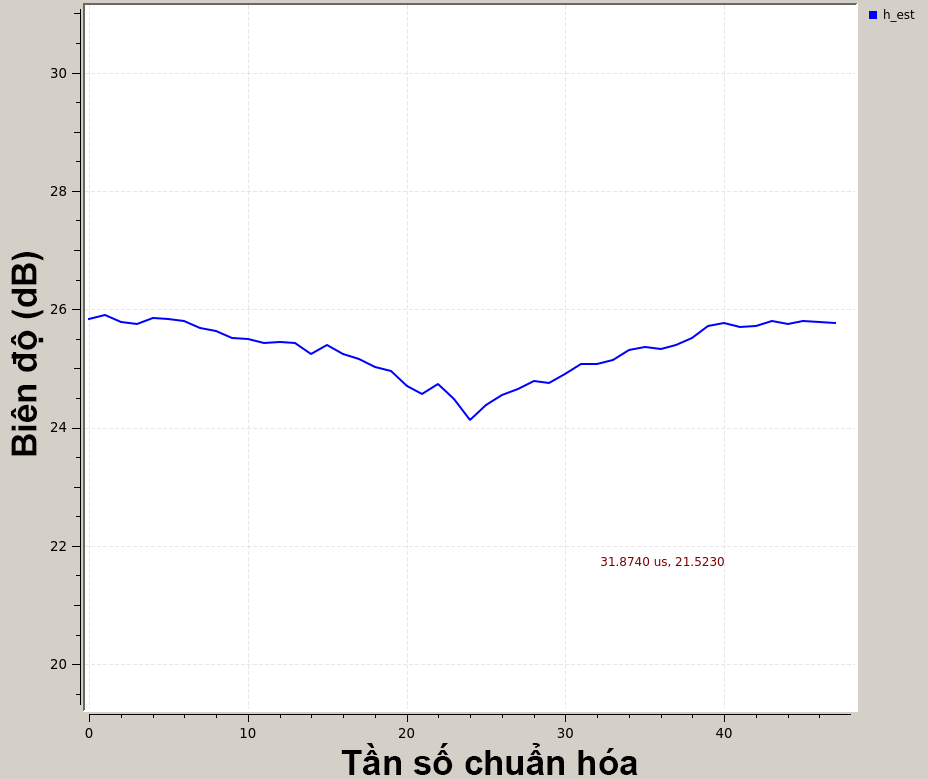}
         \caption{GNU Radio: (trái) kênh mô phỏng và (phải) kênh ước lượng}
         \label{fig:simgnu}
     \end{subfigure}
        \caption{Kết quả mô phỏng ước lượng kênh truyền trên (a) Matlab và (b) GNU Radio.}
        \label{fig:sim}
\end{figure}

Hình~\ref{fig:sim} biểu diễn đáp ứng tần số của kênh truyền đã cho và kênh truyền ước lượng được với 3a sử dụng Matlab và 3b sử dụng GNURadio. Kênh truyền được mô phỏng theo công thức~(\ref{eq:fir}) có đáp ứng biên độ tại các tần số chuẩn hóa từ 20 đến 30 (tương ứng là sóng mang con thứ 20 đến 30) suy hao đáng kể và được bộ cân bằng ZF ước lượng khá chính xác (hình~\ref{fig:simmatlab}). Với GNU Radio và SDR4All như trên hình~\ref{fig:simgnu}, ở SNR lớn, khoảng 30~dB, về hình dạng của đáp ứng kênh truyền vẫn được đảm bảo nhưng biên độ có sự sai khác khá lớn. Điều này xảy ra do bản thân mã nguồn SDR4All đã thêm khối chuẩn hóa năng lượng (``\texttt{Power~Scaling}'') trước khi thực hiện ước lượng kênh truyền.

\begin{figure}
    \centering
    \includegraphics[width=\linewidth]{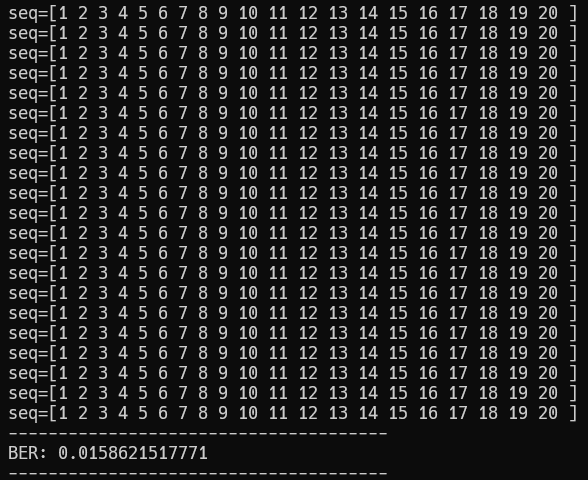}
    \caption{Chuỗi nhận được tại bên thu và tỷ số BER sau mỗi 100.000 bit thu được.}
    \label{fig:terminal}
\end{figure}

Hình~\ref{fig:terminal} biểu diễn chuỗi dữ liệu thu được trong mô phỏng dùng GNU Radio. Kết quả khá chính xác và BER được tính toán và hiển thị sau mỗi 100.000 bit.

\begin{figure}
    \centering
    \includegraphics[width=\linewidth]{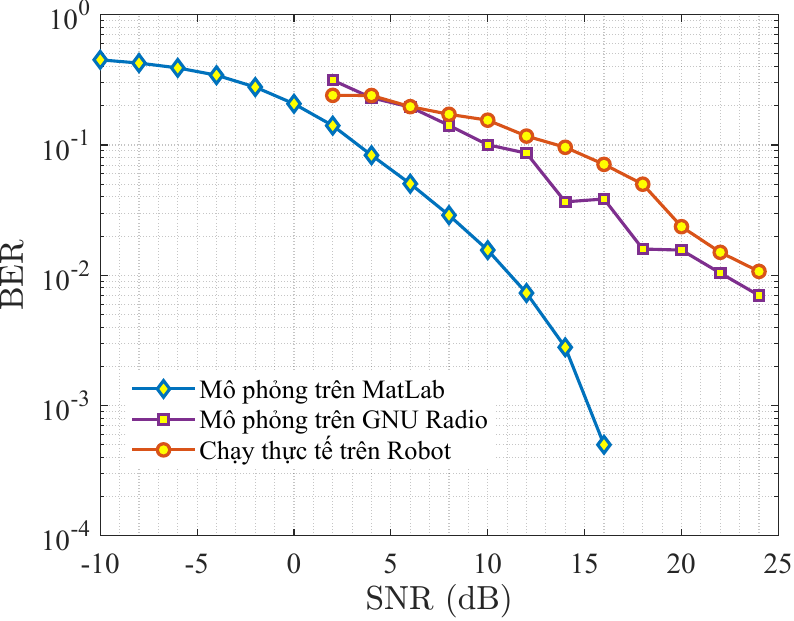}
    \caption{BER thu được thông qua mô phỏng Matlab, GNU Radio, và chạy thực tế trên BladeRF.}
    \label{fig:perf1}
\end{figure}

Hiệu năng của hệ thống được xem xét thông qua BER theo các mức SNR như trên hình~\ref{fig:perf1}. Dễ nhận thấy, với mô phỏng Matlab dựa trên lý thuyết, đường BER có thể xuất phát từ các mức SNR rất thấp (-10~dB) và giảm nhỏ hơn $10^{-3}$ tại SNR ngưỡng 16~dB. Mô phỏng GNU Radio sử dụng SDR4All gần với thực tế hơn, phải đến khi SNR lớn hơn 2~dB, bên thu mới nhận dạng và tách được tín hiệu của bên phát khỏi nền nhiễu. Xét về độ chính xác, rõ ràng SDR4All thấp hơn đáng kể so với lý thuyết và chỉ đạt tiệm cận BER$~\approx~10^{-2}$ tại các ngưỡng SNR lớn hơn 20~dB. Cần lưu ý rằng, các giá trị BER thu được từ SDR4All khá cao do chưa bao gồm các khối mã sửa lỗi.

\subsection{Kết quả thực nghiệm}

\subsubsection{Thiết lập thí nghiệm}

\begin{figure}
    \centering
    \includegraphics[width=\linewidth]{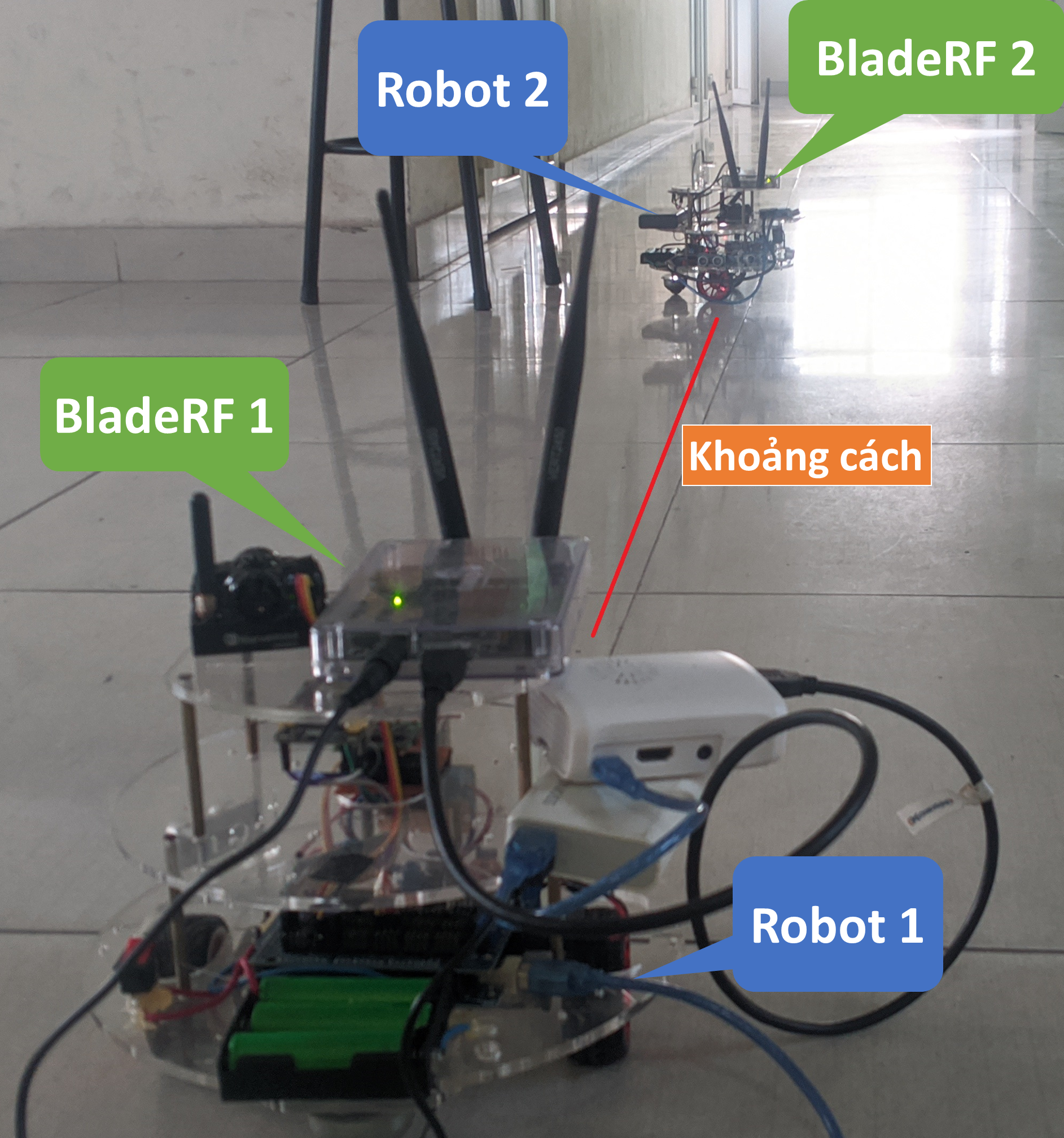}
    \caption{Hình ảnh hệ thống thực nghiệm gồm hai robot và hai BladeRF.}
    \label{fig:realsys}
\end{figure}

Hai robot cỡ nhỏ sử dụng BladeRF để truyền tín hiệu được thiết lập như trên hình~\ref{fig:realsys}. Các thông số chạy thực nghiệm được cho trên bảng~\ref{tab:real_param}. Địa điểm thực nghiệm là tầng 7 nhà E3, Trường Đại học Công nghệ - ĐHQGHN, bố trí hệ thống ở ba vị trí khác nhau để đưa ra kết quả trung bình của việc ước lượng kênh truyền. Các giá trị BER vẫn được tính trung bình và thu thập sau mỗi 100.000 bit được giải điều chế. 
\begin{table}
\centering
\caption{Các tham số của hệ thống thực nghiệm ước lượng kênh truyền cho robot sử dụng SDR}
\label{tab:real_param}
\resizebox{\linewidth}{!}{%
\begin{tabular}{>{\hspace{0pt}}m{0.521\linewidth}|>{\hspace{0pt}}m{0.408\linewidth}} 
\hline\hline
\multicolumn{1}{>{\centering\hspace{0pt}}m{0.521\linewidth}|}{\textbf{Thông số}} & \multicolumn{1}{>{\centering\arraybackslash\hspace{0pt}}m{0.408\linewidth}}{\textbf{Giá trị}} \\ 
\hline
Tần số sóng mang & $f_c = 2,515$~GHz \\ 
\hline
Loại điều chế & BPSK\par{}(Binary phase shift keying) \\ 
\hline
Chuỗi dữ liệu truyền đi & [$1, 2,~\ldots, 20$] \\ 
\hline
Độ dài FFT & $N_{fft} = 128$ \\ 
\hline
Số sóng mang con được sử dụng & $N_{Occ} = 48$ \\ 
\hline
Kiểu chèn Pilot & Block-type \\ 
\hline
Đồ dài tiền tố vòng & $N_{CP} = 16$ \\
\hline
\end{tabular}
}
\end{table}

\subsubsection{Kết quả}

Độ chính xác của việc ước lượng kênh truyền khi sử dụng các thiết bị BladeRF thực so với mô phỏng trên Matlab và GNU Radio được biểu diễn trên hình~\ref{fig:perf1} trong trường hợp hai robot đứng yên không có vật cản ở giữa. Cả ba trường hợp mô phỏng và thực nghiệm đều cho BER giảm khi SNR tăng lên. Tuy nhiên, BER trong trường hợp thực nghiệm thực tế lớn hơn so với mô phỏng Matlab và GNU Radio do sự phức tạp của kênh truyền thực. Tương tự như GNU Radio, các BladeRF cũng yêu cầu SNR đạt đến một ngưỡng SNR lớn nhất định để phát hiện và tách được tín hiệu nguồn khỏi nhiễu, theo thực nghiệm, ở khoảng 2-4~dB.
\begin{figure}
    \centering
    \includegraphics[width=\linewidth]{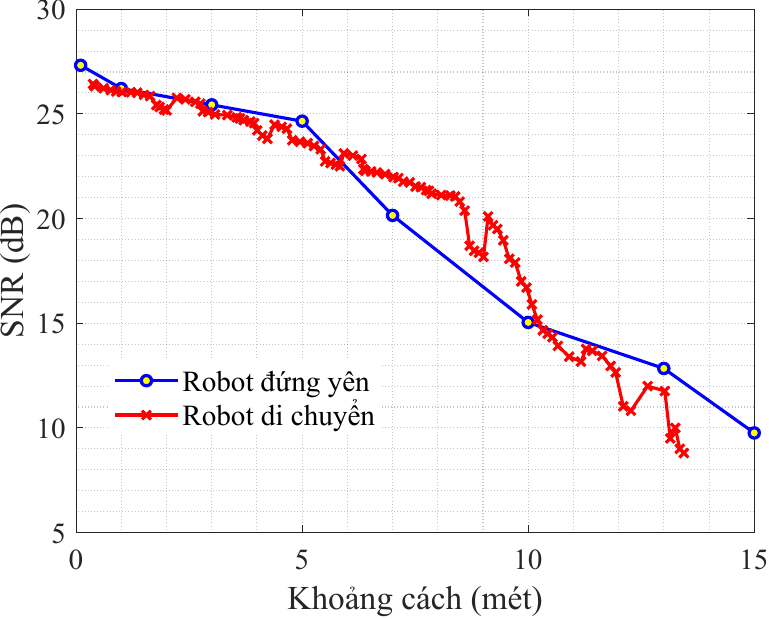}
    \caption{Sự thay đổi của SNR theo khoảng cách giữa hai robot.}
    \label{fig:disvssnr}
\end{figure}

Tiếp đến, sự phụ thuộc của SNR vào khoảng cách giữa hai robot sẽ được xem xét trong hai trường hợp: (i) robot đứng yên tại 8 khoảng cách khác nhau (0, 1, 3, 5, 7, 10, 13, và 15 mét) và (ii) robot di chuyển với vận tốc 0,3~m/s. Việc thu thập dữ liệu được thực hiện liên tục. Kết quả thu được như trên hình~\ref{fig:disvssnr} cho thấy sự suy giảm nhanh chóng của giá trị SNR khi khoảng cách giữa hai robot tăng dần. Độ dốc của đường SNR ở cả hai trạng thái đo đều khá giống nhau, dù vẫn có những sự không ổn định nhất định với dữ liệu được thu thập khi robot di chuyển. Đặc biệt, khi khoảng cách lớn hơn 13 mét, robot thu không ghi nhận được giá trị SNR hay robot thu không phát hiện ra tín hiệu bên phát ẩn trong nền nhiễu.
\begin{figure}
    \centering
    \includegraphics[width=\linewidth]{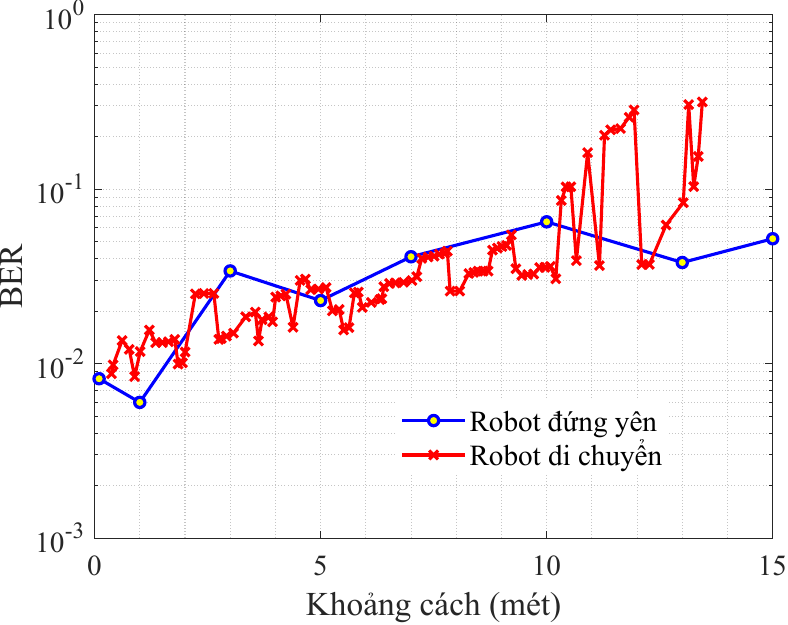}
    \caption{Sự thay đổi của BER theo khoảng cách giữa hai robot.}
    \label{fig:disvsber}
\end{figure}

 Kết quả thu được trên hình~\ref{fig:disvsber} cho thấy BER tăng dần khi khoảng cách giữa hai robot tăng. Ở trạng thái robot tĩnh, với khoảng cách 1 mét và 15 mét, BER đạt 6 * $10^{-3}$ và 5 * $10^{-2}$ tương ứng. Đây là chênh lệch lớn, tỷ lệ lỗi bit tăng lên hơn 8 lần khi khoảng cách tăng từ 1 lên 15 mét. Khi robot di chuyển, ở các khoảng cách nhỏ hơn 10 mét, BER vẫn bám khá sát với sai số khi robot tĩnh. Tuy nhiên với các khoảng cách lớn hơn, độ không ổn định tăng, dẫn đến BER có thể tăng đến 0,3 hay thậm chí không thể nhận dạng tín hiệu phát ở khoảng cách 13 mét trở lên.
\begin{figure}
    \centering
    \includegraphics[width=\linewidth]{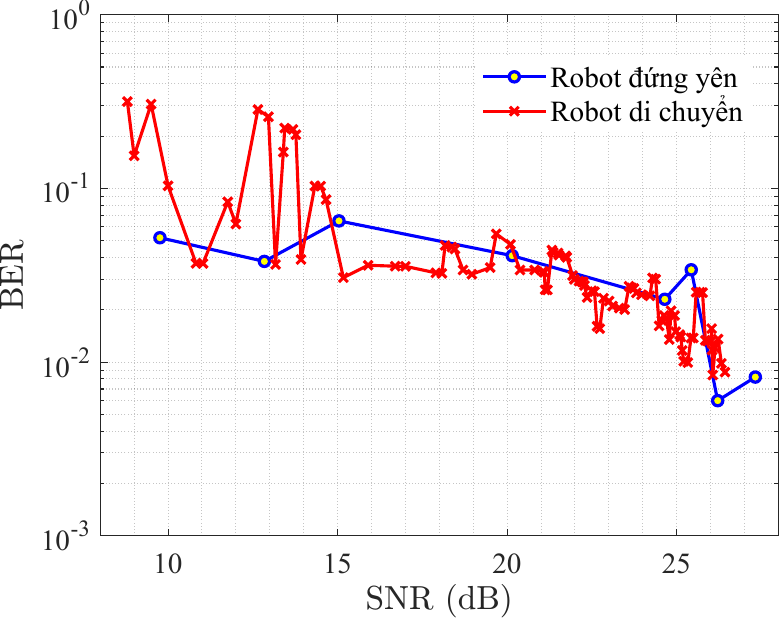}
    \caption{Sự thay đổi của BER theo tỷ lệ SNR.}
    \label{fig:snrvsber}
\end{figure}

 Hình~\ref{fig:snrvsber} biểu diễn sự phụ thuộc của BER vào SNR. Có thể thấy rằng, với mã nguồn SDR4All, tỷ lệ SNR nên ở ngưỡng 10~dB trở lên để hệ thống có thể hoạt động. Kết quả thực nghiệm này có thể gợi ý cho việc lựa chọn dải truyền thông an toàn trong hệ thống đa robot và các ngưỡng tham chiếu cho các nghiên cứu tiếp theo về điều chế thích nghi để cải thiện hiệu suất của mã nguồn SDR4All.

\section{THẢO LUẬN}
\label{Sec:dis}

Trong nghiên cứu này, điểm hạn chế cần được thảo luận đó là tỷ lệ lỗi bit khi mô phỏng trên GNU Radio cũng như chạy thực nghiệm còn cách khá xa so với mô phỏng trên Matlab. Sai số này bản thân xuất phát từ mã nguồn SDR4All. Cụ thể, như trong hình~\ref{fig:simgnu}, do việc cân bằng tín hiệu diễn ra sau khi tín hiệu đã được chuẩn hóa năng lượng, dẫn đến sai số trong việc ước lượng kênh truyền. Để tỷ số BER giảm, trong nghiên cứu~\cite{sdrnc} của chúng tôi, hai hướng giải quyết được đưa ra để đảm bảo hệ thống hoạt động tốt hơn trong các nhiệm vụ yêu cầu độ chính xác cao hơn như truyền hình ảnh: (i), thêm các khối mã kênh; (ii), chia nhỏ bản tin thành các gói tin (packet) nhỏ hơn, thêm cơ chế kiểm tra lỗi cho từng gói tin, bổ sung cơ chế phản hồi cho bên thu để thông báo cho bên phát truyền lại các gói tin bị lỗi.

\section{KẾT LUẬN}
\label{Sec:KetLuan}
Nhóm nghiên cứu đã xây dựng mô hình thực nghiệm hệ thống truyền thông đa robot sử dụng SDR. Một số kịch bản mô phỏng và thực nghiệm được đưa ra để kiểm chứng khả năng ước lượng kênh truyền của hai robot khi đứng yên và di chuyển liên tục. Kết quả được đưa ra trong bài báo chứng minh tính khả thi của việc ứng dụng các thiết bị SDR vào các nghiên cứu về mạng truyền thông đa robot trong tương lai cũng như bộc lộ một số điểm hạn chế của bộ nhận dạng trong mã nguồn SDR4All. Trong tương lai, nhóm nghiên cứu sẽ xem xét việc tăng thêm số robot hoạt động trong mạng cũng như phát triển thêm các bộ cân bằng kênh khác cho mã nguồn SDR4All phục vụ cho các mục đích nghiên cứu và giảng dạy.

Các lưu đồ trên GNU Radio đã được sử dụng trong bài báo, các khối do nhóm nghiên cứu xây dựng có thể tham khảo tại:

\noindent \url{https://github.com/DoHaiSon/SDR_NC/tree/BER}

\section*{LỜI CẢM ƠN}
Nghiên cứu này được tiến hành trong khuôn khổ đề tài 
QG.21.26 ``Giải pháp chống, chịu nhiễu duy trì mạng đa robot trong môi trường động" của ĐHQGHN.

\bibliographystyle{IEEEtran}
\balance
\bibliography{reference}
\end{document}